\pdfoutput=1

\documentclass[11pt,a4paper]{article}

\usepackage{bibunits}
\usepackage{placeins}
\usepackage{authblk}

\usepackage[hang,flushmargin]{footmisc}
\usepackage[hyperref]{acl}

\usepackage{times} 
\usepackage{latexsym}
\usepackage{textcomp}

\usepackage{caption}  

\usepackage[latin1]{inputenc}
\usepackage[T1]{fontenc}
\usepackage{microtype} 
\usepackage{url}
\usepackage{booktabs}
\usepackage{tabularx}
\usepackage{nicefrac}

\usepackage{amsmath}
\DeclareMathOperator*{\argmax}{arg\!max}
\DeclareMathOperator*{\argmin}{arg\!min}

\newcommand{\Ni}{(1)~}
\newcommand{\Nii}{(2)~} 
\newcommand{\Niii}{(3)~}
\newcommand{\Niv}{(4)~}
\newcommand{\Nv}{(5)~}

\usepackage{mathtools}
\DeclarePairedDelimiterX{\kldivx}[2]{(}{)}{%
  #1\;\delimsize\|\;#2%
}
\newcommand{\kldiv}{\mathrm{KL}\kldivx}

\usepackage{graphicx}
\graphicspath{{acl22-fine-tuning-for-active-learning-figures/}}
\DeclareGraphicsExtensions{.pdf}
\setkeys{Gin}{pagebox=artbox}

\usepackage{stfloats}

\usepackage{multirow}
\usepackage{siunitx}
\newcolumntype{P}{p{0.3cm}}

\begin{document}

\defaultbibliography{acl22-fine-tuning-for-active-learning-lit}
\defaultbibliographystyle{acl_natbib}

\begin{bibunit}
\title{Revisiting Uncertainty-based Query Strategies for\\ Active Learning with Transformers}

\date{}

\author{\textbf{Christopher Schr{\"o}der}}
\author{\textbf{Andreas Niekler}}
\author{\textbf{Martin Potthast}}

\affil{Leipzig University\vspace*{-0.8em}}

\maketitle

\begin{abstract}
Active learning is the iterative construction of a classification model through targeted labeling, enabling significant labeling cost savings. As most research on active learning has been carried out before transformer-based language models (``transformers'') became popular, despite its practical importance, comparably few papers have investigated how transformers can be combined with active learning to date. This can be attributed to the fact that using state-of-the-art query strategies for transformers induces a prohibitive runtime overhead, which effectively nullifies, or even outweighs the desired cost savings. For this reason, we revisit uncertainty-based query strategies, which had been largely outperformed before, but are particularly suited in the context of fine-tuning transformers. In an extensive evaluation, we connect transformers to experiments from previous research, assessing their performance on five widely used text classification benchmarks. For active learning with transformers, several other uncertainty-based approaches outperform the well-known prediction entropy query strategy, thereby challenging its status as most popular uncertainty baseline in active learning for text classification.
\end{abstract}

\section{Introduction}

Collecting labeled data for machine learning can be costly and time-consuming. A key technique to minimize labeling costs has been active learning, where an oracle (e.g., a human expert) is queried to label problem instances selected that are deemed to be most informative to the learning algorithm's next iteration according to a query strategy.
 
Active learning is characterized by the real-world machine learning scenario in which large amounts of training data are unavailable, which may explain why comparably little research has investigated deep learning in this context. The recent widely successful transformer-based language models can circumvent the limitations imposed by small training datasets \citep{vaswani:2017,devlin:2019}. Pre-trained on large amounts of unlabeled text, they can be fine-tuned to a given task using far less training data than when trained from scratch. However, their high number of model parameters renders them computationally highly expensive, for query strategies that are targeted at neural networks or text classification \citep{settles:2007,zhang:2017}, resulting in prohibitive turnaround times between labeling steps.

In this paper, we systematically investigate uncertainty-based query strategies as a computationally inexpensive alternative. Despite their relative disadvantages in traditional active learning, when paired with transformers, they are highly effective as well as efficient. Our extensive experiments assess a multitude of combinations including state-of-the-art transformer models BERT \citep{devlin:2019} and DistilRoBERTa \citep{sanh:2019}, five well-known sentence classification benchmarks, and five query strategies.%
\footnote{Code: {\url{https://github.com/webis-de/ACL-22}}}

\section{Related Work}

\enlargethispage{\baselineskip}
Uncertainty-based query strategies used to be the most common choice in active learning, using uncertainty scores obtained from the learning algorithm \citep{lewis:1994}, estimates obtained via ensembles \citep{krogh:1994,raychaudhuri:1995}, or prediction entropy \citep{holub:2008}. More recently---predating transformers---neural network-based active learning predominantly employed query strategies that select problem instances according to
\Ni
the magnitude of their backpropagation-induced gradients \citep{settles:2007,zhang:2017}, where instances causing a high-magnitude gradient inform the model better, and
\Nii
representativity-based criteria (e.g., coresets \citep{sener:2018}), which select instances from a vector space to geometrically represent the full dataset.

\noindent For today's deep neural networks, ensembles are too computationally expensive, and prediction entropy has been observed to be overconfident \citep{guo:2017,lakshminarayanan:2017}. The exception are flat architectures, where, among others, \citet{prabhu:2019} showed fastText \citep{joulin:2017} to be effective,  well-calibrated, and computationally efficient. Prior to transformers, query strategies relying on expected gradient length \citep{settles:2007} achieved the best results on many active learning benchmarks for text classification \citep{zhang:2017}. Gradients depend on the current model, which means, when used for a query strategy, they scale with the vast number of a transformer's parameters, and moreover, they need to be computed per-instance instead of batch-wise, thereby becoming computationally expensive.

The cost of ensembles, the adverse scaling of network parameters in gradient-based strategies, and a history of deeming neural networks to be overconfident effectively rule out the most predominantly used query strategies. This might explain why transformers, despite the success of fine-tuning them for text classification \citep{howard:2018,yang:2019,sun:2020}, have only very recently been considered at all in combination with active learning \citep{lu:2020,yuan:2020,ein-dor:2020,margatina:2021b}. All of the related works mitigate the computationally complex query strategies by subsampling the unlabeled data before querying \citep{lu:2020,ein-dor:2020,margatina:2021b}, by performing fewer queries with larger sample sizes \citep{yuan:2020,margatina:2021b}, or by tailoring to less expensive settings, namely binary classification \citep{ein-dor:2020}. Subsampling, however, introduces additional randomness which can aggravate comparability across experiments, and large sample sizes increase the amount of labeled data, which is contrary to  minimizing the labeling effort.

Due to this computationally challenging setting, the uncertainty-based prediction entropy query strategy \citep{roy:2001,schohn:2000} is therefore a frequently used baseline and a lowest common denominator in recent work on active learning for text classification \citep{zhang:2017,lowell:2019,prabhu:2019,ein-dor:2020,lu:2020,yuan:2020,margatina:2021b,zhang:2021}. Apart from being employed as baselines, uncertainty-based query strategies have not been systematically analyzed in conjunction with transformers, and moreover, comparisons to the previous benchmarks by \citet{zhang:2017} have been omitted by the aforementioned related work. Our work not only closes this gap, but also reevaluates the relative strength of uncertainty-based approaches, including two recently largely neglected strategies, thereby challenging the status of prediction entropy as the most popular baseline.

\section{Transformer-based Active Learning}
\label{sec:transformer-based-active-learning}

The goal of active learning is to minimize the labeling costs of training data acquisition while maximizing a model's performance (increase) with each newly labeled problem instance. In contrast to regular supervised text classification (``passive learning''), it operates  iteratively, where in each iteration
\Ni
a so-called query strategy selects new instances for labeling according to an estimation of their informativeness,
\Nii
an oracle (e.g., a human expert) provides the respective label, and 
\Niii
a learning algorithm either uses the newly labeled instance for its next learning step, or a model is retrained from scratch using all previously labeled instances. This work considers pool-based active learning \citep{lewis:1994}, where the query strategies have access to all unlabeled data. Notation-wise, we denote instances by $x_1, x_2, \dots, x_n$, the number of classes by $c$, the respective label for instance $x_i$ by $y_i$ (where $\forall i: y_i \in \{1, \dots, c\}$), and $P(y_i|x_i)$ is a probability-like predicted class distribution.

\paragraph{Query Strategies} 

{
\begingroup
\setlength{\abovedisplayskip}{5pt}
\setlength{\belowdisplayskip}{5pt}
\setlength{\abovedisplayshortskip}{5pt}
\setlength{\belowdisplayshortskip}{5pt}
    
We consider three well-known uncertainty-based query strategies, one recent state-of-the-art strategy that coincidentally also includes uncertainty, and a random baseline:\\
\Ni
Prediction Entropy (PE; \citealp{roy:2001,schohn:2000}) selects instances with the highest entropy in the predicted label distribution with the aim to reduce overall entropy:
\begin{equation*}
\argmax_{x_i}~\left[-\sum_{j=1}^{c} P(y_i = j|x_i)\log{}P(y_i = j|x_i)~\right]
\end{equation*}
\Nii
Breaking Ties (BT; \citealp{scheffer:2001,luo:2005}) takes instances with the minimum margin between the top two most likely probabilities: 
\begin{equation*}
\argmin_{x_i}~\Big[P(y_i = k_1^{*}|x_i) - P(y_i = k_2^{*}|x_i)\Big]
\end{equation*}
where $k_1^{*}$ is the most likely label in the posterior class distribution $P(y_i|x_i)$, and $k_2^{*}$ the second most likely label respectively. {\em In the binary case}, this margin is small iff the label entropy is high, which is why BT and PE then select the same instances.\\
\Niii
Least Confidence (LC; \citealp{culotta:2005}) selects instances whose most likely label has the least confidence according to the current model:
\begin{equation*}
\argmax_{x_i}~\Big[1 - P(y_i = k_1^{*}|x_i)\Big]
\end{equation*}
\Niv
Contrastive Active Learning (CA; \citealp{margatina:2021b})  selects instances with the maximum mean Kullback-Leibler~(KL) divergence between the predicted class distributions (``probabilities'') of an instance and each of its $m$ nearest neighbors:
\begin{equation*}
\argmax_{x_i}~\left[~\frac{1}{m}\sum\limits_{j=1}^{m}\kldiv{P(y_j|x_j^{knn})}{P(y_i|x_i)}~\right]
\end{equation*}
where the instances $x_j^{knn}$ are the $m$ nearest neighbors of instance $x_i$.\\
\Nv
Random Sampling (RS), a commonly used baseline, draws uniformly from the unlabeled pool.

\endgroup
}

\paragraph{Oracle}

The oracle is usually operationalized using the training datasets of existing benchmarks: To ensure comparability with the literature, we pick important standard text classification tasks.

\paragraph{Classification}
 
We fine-tune BERT \citep{devlin:2019} and DistilRoBERTa \citep{sanh:2019} on several natural language understanding datasets. BERT is well-researched as transformer and has recently also shown strong results in active learning \citep{yuan:2020,ein-dor:2020,margatina:2021b}. The model consists of 24 layers, hidden units of size 1024 and 336M~parameters in total. DistilRoBERTa, by contrast, is a more parameter-efficient alternative which has merely six layers, hidden units of size~768, and 82M~parameters.
We also trained a passive model on the full data.
  
\begin{table}[t!]%
\centering
\fontsize{9pt}{10pt}\selectfont%
\renewcommand{\tabcolsep}{2.5pt}%
\begin{tabular}[t]{@{}l@{\hspace{7pt}}ccrr@{}}
\toprule
\bfseries Dataset Name {\tiny (ID)} & \bfseries Type & \bfseries Classes & \bfseries Training & \bfseries Test \\
\midrule
AG's News {\scriptsize (AGN)}       & N & 4 &  120,000 & \textsuperscript{(*)} 7,600 \\
Customer Reviews {\scriptsize (CR)} & S   & 2 &    3,397 &                         378 \\
Movie Reviews {\scriptsize (MR)}    & S   & 2 &    9,596 &                       1,066 \\
Subjectivity {\scriptsize (SUBJ)}   & S   & 2 &    9,000 &                       1,000 \\
TREC-6 {\scriptsize (TREC-6)}       & Q   & 6 &    5,500 & \textsuperscript{(*)} 500 \\
\bottomrule
\end{tabular}
\caption{Key information about the examined datasets. The dataset type was abbreviated as follows: N: News, S: Sentiment, Q: Questions. (*):~Predefined test sets were available and adopted.}
\label{table-datasets}
\end{table}

The classification model consists of the respective transformer, on top of which we add a fully connected projection layer, and a final softmax output layer. 
We use the ``[CLS]'' token that is computed by the transformer as sentence representation. 
Regarding fine-tuning, we adopt the combination of discriminative fine-tuning and slanted triangular learning rates \citep{howard:2018}. The main active learning routine is then as follows:
\Ni
The query strategy, either using the model from the previous iteration, or sampling randomly, selects 25~instances from the unlabeled pool.
\Nii
The oracle provides labels for these instances.
\Niii
The next model is trained using all data labeled so far.

\paragraph{Baselines}
For comparison, we consider a linear SVM, and KimCNN \citep{kim:2014}, which have been used extensively in text classification, disregarding active learning. We adopted the KimCNN parameters from \citet{kim:2014} and \citet{zhang:2017}.

\section{Evaluation}

\begin{table}[!t]%
\centering
\fontsize{9pt}{10pt}\selectfont%
\renewcommand{\tabcolsep}{4.5pt}%
\begin{tabular}[t]{@{}l@{\hspace{4mm}}l@{\hspace{5mm}}rrrr@{}}
\toprule
\multirow{2}{*}[-0.2em]{{\bfseries Model}} & \multirow{2}{*}[-0.2em]{\bfseries Strategy} & \multicolumn{2}{@{}r@{}}{\bfseries Mean Rank\hspace*{2mm}} & \multicolumn{2}{@{}r@{}}{\bfseries Mean Result}\\
\cmidrule(r){3-4}\cmidrule(l){5-6} & & \bfseries Acc. & \bfseries AUC & \bfseries Acc. & \bfseries AUC\\
\midrule
SVM & PE  & 1.80 & 2.60 & 0.764 & 0.663 \\
 & BT  & \bfseries 1.60 & \bfseries 1.60 & \bfseries 0.767 & \bfseries 0.697 \\
 & LC  & 3.00 & 2.60 & 0.751 & 0.672 \\
 & CA  & 5.00 & 5.00 & 0.667 & 0.593 \\
 & RS  & 3.00 & 2.60 & 0.757 & 0.686 \\
\midrule
KimCNN & PE  & 1.60 & 2.40 & \bfseries 0.818 & 0.742 \\
 & BT  & \bfseries 1.60 & \bfseries 2.00 & \bfseries 0.818 & \bfseries 0.750 \\
 & LC  & 3.80 & 2.80 & 0.810 & 0.732 \\
 & CA  & 3.80 & 4.80 & 0.793 & 0.711 \\
 & RS  & 3.60 & 2.40 & 0.804 & 0.749 \\
\midrule
D.RoBERTa & PE  & 2.60 & 3.00 & 0.901 & 0.856 \\
 & BT  & 2.20 & \bfseries 1.80 & 0.902 & \bfseries 0.864 \\
 & LC  & \bfseries 1.40 & 2.00 & \bfseries 0.904 & 0.860 \\
 & CA  & 3.00 & 3.40 & 0.901 & 0.852 \\
 & RS  & 5.00 & 4.20 & 0.884 & 0.853 \\
\midrule
BERT & PE  & 2.40 & 2.40 & 0.909 & 0.859 \\
 & BT  & \bfseries 2.00 & \bfseries 1.60 & 0.914 & \bfseries 0.873 \\
 & LC  & 2.20 & 3.80 & \bfseries 0.917 & 0.866 \\
 & CA  & 2.80 & 2.60 & 0.916 & 0.872 \\
 & RS  & 5.00 & 4.00 & 0.899 & 0.861 \\
\bottomrule
\end{tabular}
\caption{The ``Mean Rank'' columns show the mean rank when ordered by mean accuracy (Acc.) after the final iteration and by overall AUC. The ``Mean Result'' columns show the mean accuracy and AUC.}
\label{table-results-summary}
\end{table} 

We evaluate five query strategies in combination with BERT,  {DistilRoBERTa} and two baselines.

\paragraph{Datasets and Experimental Setup}

In Table~\ref{table-datasets}, we show the five datasets employed, which have previously been used to evaluate active learning: AG's News (AGN; \citealp{zhang:2015}), Customer Reviews (CR; \citealp{hu:2004}), Movie Reviews (MR; \citealp{pang:2005}), Subjectivity (SUBJ; \citealp{pang:2004}), and {TREC-6} \citep{li:2002}. These datasets encompass binary and multi-class classification in different domains, and they are class-balanced, except for {TREC-6}. Where available, we employed the pre-existing test sets, or otherwise a random sample of~10\%.

We follow the experiment setup of \citet{zhang:2017}: 25~training instances are used to train the first model, followed by 20~active learning iterations, during each of which 25~instances are queried and labeled. Using 10\%~of the so far labeled data as validation set, we stop early \citep{duong:2018} when accuracy surpasses~98\%, or the validation loss does not increase for five epochs.

\begin{figure*}[t]
\centering
\includegraphics[width=0.77\textwidth]{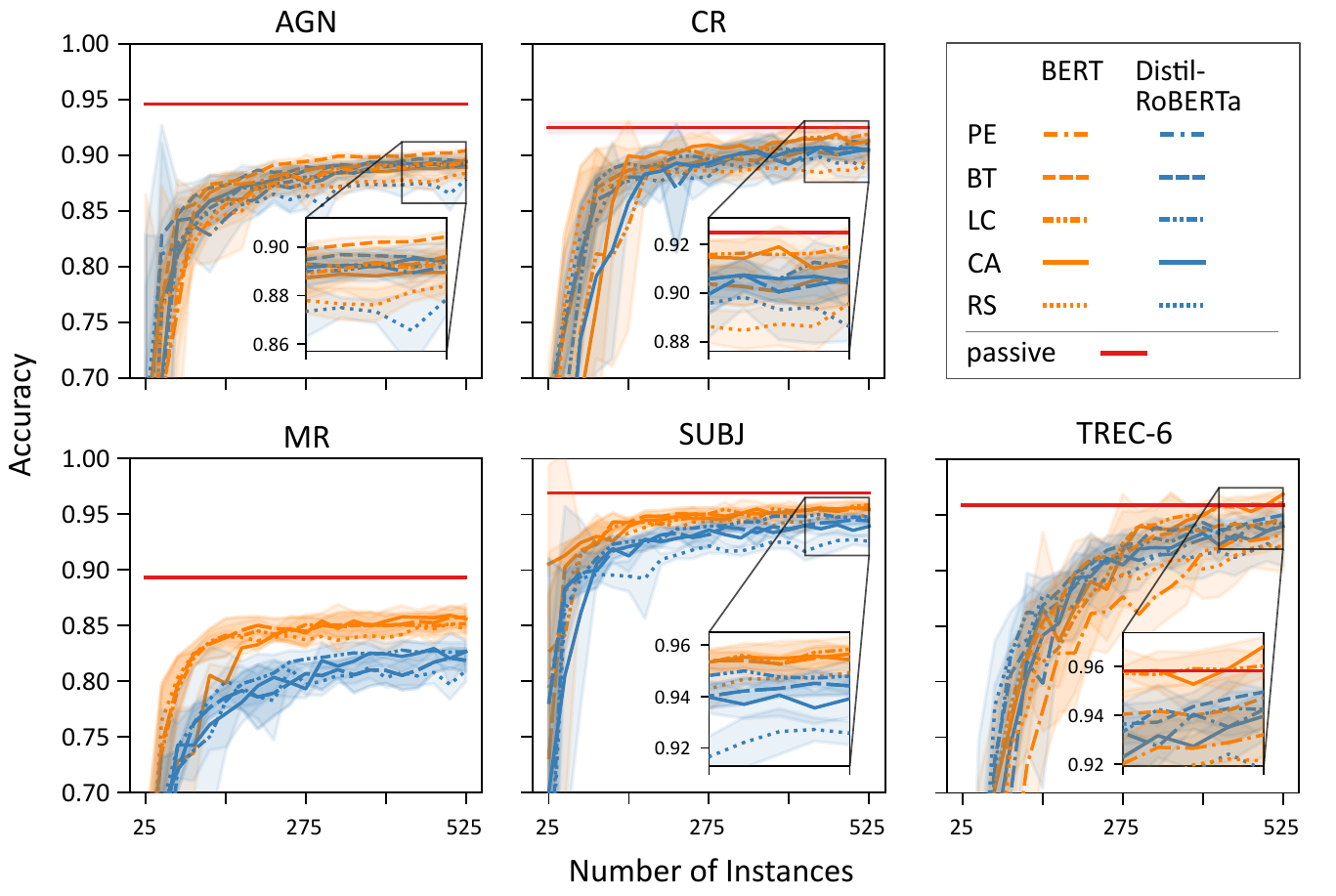}
\caption{Active learning curves of BERT and DistilRoBERTa when combined with five query strategies: Prediction Entropy (PE), Breaking Ties (BT), Least Confidence (LC), Contrastive Active Learning (CA), and Random Sampling (RS). The tubes around the lines represent standard deviation over five runs. For comparison, the horizontal line depicts a passive text classification for which BERT has been trained using the entire training set.}
\label{figure-learning-curves}
\end{figure*} 

\paragraph{Results}

For each combination of dataset, model, and query strategy, Figure~\ref{figure-learning-curves} shows the respective learning curves. The horizontal line shows the best model's score when trained on the full dataset, which four out of five datasets approach very closely, or even exceed. As expected, BERT generally achieves steeper learning curves than DistilRoBERTa, but surprisingly, during later iterations DistilRoBERTa reaches scores only slightly worse than BERT for all datasets except MR. Regarding query strategies, RS is a strong contender during early iterations, e.g., as can be seen for the first few iterations of~CR. This is partly because all but one of the datasets are balanced, but nevertheless, RS is eventually outperformed by the other strategies in most cases. For imbalanced datasets, \citet{ein-dor:2020} have shown RS to be less effective, which we can confirm for \mbox{TREC-6}. 
\begin{table}[t!]%
\centering
\fontsize{9pt}{10pt}\selectfont%
\renewcommand{\tabcolsep}{3pt}%}
{
\renewcommand{\arraystretch}{1.03}%
\begin{tabular}[t]{@{}l p{2cm}@{\ \ }lcr@{}}
\toprule
\bfseries Dataset & \bfseries Model & \bfseries Strategy & \bfseries Acc. & \bfseries Data Use\\
\midrule
\multirow{3}{*}{AGN} & BERT  & BT& 0.904 & 0.4\%\\
& BERT  & passive (ours) & 0.946 & 100.00\%\\
& XLNet$^1$                 & passive & 0.955 & 100.00\%\\
\midrule
\multirow{3}{*}{CR} & BERT       & LC & 0.919 & 15.45\%\\
& BERT                 & passive (ours) & 0.925 & 100.00\%\\
& HAC$^2$          & passive & 0.889 & 100.00\%\\
\midrule
\multirow{3}{*}{MR} & BERT     & PE, BT & 0.857 & 0.547\% \\
& BERT                 & passive (ours) & 0.893 & 100.00\%\\
& SimCSE$^3$                  & passive & 0.884 & 100.00\%\\
\midrule                   
\multirow{3}{*}{SUBJ} & BERT       & LC & 0.958 & 5.83\% \\
& BERT                  & passive (ours) & 0.969 & 100.00\%\\
& AdaSent$^4$               & passive & 0.955 & 100.00\%\\
\midrule                   
\multirow{3}{*}{TREC-6} & BERT         & CA & 0.968 & 9.55\% \\
& BERT & passive (ours) & 0.958 & 100.00\%\\
& RCNN$^5$                  & passive & 0.962 & 100.00\%\\
\bottomrule
\end{tabular}%
}
\caption{%
Best final accuracy compared to (our) passive classification and state-of-the-art text classification: $^1$\citet{yang:2019}, $^2$\citet{zheng:2019}, $^3$\citet{gao:2021}, $^4$\citet{zhao:2015}, $^5$\citet{tay:2018}. ``Data Use'' indicates proportion of training data used.}
\label{table-comparison-acc}
\end{table}
While in terms of area under the learning curve (AUC) there seems to be no overall best strategy, PE/BT and CA often show very steep learning curves. 

In Table~\ref{table-results-summary}, we rank the query strategies by their average accuracy and AUC results, ranging from 1 (best) to 5 (worst). We also report their average accuracy and AUC per model and query strategy. Surprisingly, we can see that PE, a commonly used and proven to be strong baseline, which has been a lowest common denominator in recent work on active learning for text classification \citep{zhang:2017,lowell:2019,prabhu:2019,ein-dor:2020,lu:2020,yuan:2020,margatina:2021b,zhang:2021}, is on average outranked by BT when using transformers. BT achieves the best AUC ranks and scores, and in many cases also the best accuracy ranks and scores. It seems to be similarly effective on the baselines as well. Moreover, LC also outperforms PE for DistilRoBERTa where it even competes with BT. Detailed accuracy and AUC scores including standard deviations are reported in Appendix Tables \ref{table-results-acc} \& \ref{table-results-auc}.

Table~\ref{table-comparison-acc} compares the best model trained via active learning per dataset against passive text classification, namely 
\Ni
our own model trained on the full training set, and
\Nii
state-of-the-art results. The largest discrepancy between active learning and passive text classification is observed on AGN, which is also the largest dataset from which the active learning models use less than~1\% for training. Otherwise, all models are close to or even surpass the state of the art, using only between~0.4\% and~14\% of the data. Noteworthy, LC achieves the best accuracy result for two datasets, while the strong baseline PE and the state-of-the-art approach CA perform best on only one dataset each. 

In Table~\ref{table-comparison-auc}, we report the best AUC scores per dataset, and compare them to previous work. BT ranks highest in two out of three cases with CA achieving the best result on the remaining two datasets. BERT achieves the best AUC scores on all datasets with a considerable increase in AUC compared to \citet{zhang:2017}.

\begin{table}[t!]%
{
\centering
\fontsize{9pt}{10pt}\selectfont%
\renewcommand{\tabcolsep}{3pt}%}
\renewcommand{\arraystretch}{1.05}%
\begin{tabular}[t]{@{}l@{\hspace{15pt}}l@{\hspace{15pt}}c@{}}
\toprule
\bfseries Dataset & \bfseries Model & \bfseries AUC \\
\midrule
\multirow{2}{*}{AGN}  & BERT {\scriptsize (BT, ours)} & 0.875\\
& --\\
\midrule
\multirow{2}{*}{CR} & BERT {\scriptsize (PE, BT; ours)} & 0.877\\
& CNN$^6$              & 0.743\\
\noalign{\vskip -0.5pt}
\midrule
\multirow{2}{*}{MR} & BERT {\scriptsize (PE, BT; ours)} & 0.833 \\
& CNN$^6$              & 0.707\\
\noalign{\vskip -0.5pt}
\midrule
\multirow{2}{*}{SUBJ} & BERT {\scriptsize (CA, ours)} & 0.943\\
& CNN$^6$              & 0.856\\
\noalign{\vskip -0.35pt}
\midrule
\multirow{2}{*}{TREC-6} & BERT {\scriptsize (CA, ours)} & 0.868\\
& --\\
\bottomrule 
\end{tabular}%
\par  
}
\caption{%
Best area under curve (AUC) scores (averaged over five runs) compared to \citet{zhang:2017}.}
\label{table-comparison-auc}
\end{table}

In summary, we use recent transformer models in combination with several query strategies to evaluate a previously established but lately neglected benchmark. We find that the PE baseline is outperformed by BT, which, as a reminder, selects the same instances as PE for binary classification, but shows superior results on multi-class datasets. We conclude that BT, which even outperforms the state-of-the-art strategy CA in many cases, is therefore a strong contender to become the new default uncertainty-based baseline. Finally, DistilRoBERTa, using less than 25\% of BERT's parameters, achieves results that are remarkably close to BERT at only a fraction of the overhead. Considering the computational burdens that motivated this work, this increase in efficiency is often preferable from a practitioner's perspective.

\section{Conclusions}

An investigation of the effectiveness of uncertainty-based query strategies in combination with BERT and DistilRoBERTa for active learning on several sentence classification datasets shows that uncertainty-based strategies still perform well. We evaluate five query strategies on an established benchmark, for which we achieve results close to state-of-the-art text classification on four out of five datasets, using only a small fraction of the training data. Contrary to current literature, prediction entropy, the supposedly strongest uncertainty-based baseline, is outperformed by several uncertainty-based strategies on this benchmark---in particularly by the breaking ties strategy. This invalidates the common practice of solely relying on prediction entropy as baseline, and shows that uncertainty-based strategies demand renewed attention especially in the context of transformer-based active learning.  

\section*{Acknowledgments}

We thank the anonymous reviewers for their valu-
able and constructive feedback. This research was partially funded by the
Development Bank of Saxony (SAB) under project
number 100335729.

\section*{Ethical Considerations}

Research on active learning improves the labeling of data, by efficiently supporting the learning algorithm with targeted information, so that overall less data has to be labeled. This could contribute to creating machine learning models, which would otherwise be infeasible, either due to limited budget, or time. Active learning can be used for good or bad, and our contributions would---in both cases--show how to make this process more efficient.

Moreover, we use pre-trained models, which can contain one or more types of bias. Bias, however, affects all approaches based on fine-tuning pre-trained language models, but therefore this has to be kept in mind and mitigated all the more.

\putbib
\end{bibunit}

\begin{bibunit}
\appendix

\section*{Supplementary Material}

The experiments can be reproduced using the code that is referenced on the first page\footnote{\url{https://github.com/webis-de/ACL-22}}. In the following, we summarize important details for reproduction, including details on the results.

\begin{table*}[!b]%
\centering
\fontsize{8pt}{9pt}\selectfont%
\renewcommand{\tabcolsep}{12pt}%
\begin{tabular}{@{}ll@{\hspace{19pt}}lllll@{}}
\toprule
\textbf{Dataset} & \textbf{Model} & \multicolumn{5}{c}{\textbf{Query Strategy}}\\
\cmidrule{3-7} & & \multicolumn{1}{c}{\hspace*{-12pt}PE} & \multicolumn{1}{c}{BT} & \multicolumn{1}{c}{LC} & \multicolumn{1}{c}{CA} & \multicolumn{1}{c}{RS}\\
\midrule
\multirow{4}{*}{AGN}  & SVM & \num{0.804+-0.000} & \num{0.804+-0.000} & \num{0.802+-0.009} & \num{0.539+-0.088} & \num{0.801+-0.006}\\
 & KimCNN & \num{0.871+-0.004} & \num{0.874+-0.005} & \num{0.856+-0.012} & \num{0.814+-0.015} & \num{0.866+-0.007}\\
 & DistilRoBERTa & \num{0.892+-0.002} & \num{0.894+-0.003} & \num{0.894+-0.002} & \num{0.894+-0.008} & \num{0.879+-0.008}\\
 & BERT & \num{0.896+-0.003} & \bfseries\num{0.904+-0.002} & \num{0.894+-0.006} & \num{0.889+-0.014} & \num{0.884+-0.003}\\
\midrule
\multirow{4}{*}{CR}  & SVM & \multicolumn{2}{c}{\num{0.757+-0.000}}  & \num{0.755+-0.014} & \num{0.742+-0.022} & \num{0.763+-0.025}\\
 & KimCNN & \multicolumn{2}{c}{\num{0.765+-0.012}}  & \num{0.762+-0.012} & \num{0.748+-0.015} & \num{0.745+-0.014}\\
 & DistilRoBERTa & \multicolumn{2}{c}{\num{0.906+-0.007}}  & \num{0.911+-0.008} & \num{0.905+-0.011} & \num{0.886+-0.007}\\
 & BERT & \multicolumn{2}{c}{\num{0.904+-0.010}}  & \bfseries\num{0.919+-0.009} & \num{0.913+-0.005} & \num{0.896+-0.008}\\
\midrule
\multirow{4}{*}{MR}  & SVM & \multicolumn{2}{c}{\num{0.674+-0.000}}  & \num{0.650+-0.012} & \num{0.633+-0.014} & \num{0.641+-0.010}\\
 & KimCNN & \multicolumn{2}{c}{\num{0.719+-0.011}}  & \num{0.719+-0.017} & \num{0.726+-0.008} & \num{0.720+-0.013}\\
 & DistilRoBERTa & \multicolumn{2}{c}{\num{0.819+-0.012}}  & \num{0.826+-0.009} & \num{0.826+-0.011} & \num{0.809+-0.011}\\
 & BERT & \multicolumn{2}{c}{\bfseries\num{0.857+-0.009}}  & \num{0.852+-0.009} & \num{0.856+-0.015} & \num{0.846+-0.011}\\
\midrule
\multirow{4}{*}{SUBJ}  & SVM & \multicolumn{2}{c}{\num{0.843+-0.000}}  & \num{0.857+-0.006} & \num{0.827+-0.012} & \num{0.839+-0.012}\\
 & KimCNN & \multicolumn{2}{c}{\num{0.897+-0.004}}  & \num{0.880+-0.008} & \num{0.877+-0.010} & \num{0.896+-0.009}\\
 & DistilRoBERTa & \multicolumn{2}{c}{\num{0.944+-0.004}}  & \num{0.948+-0.008} & \num{0.939+-0.008} & \num{0.926+-0.005}\\
 & BERT & \multicolumn{2}{c}{\num{0.957+-0.004}}  & \bfseries\num{0.958+-0.005} & \num{0.954+-0.005} & \num{0.949+-0.003}\\
\midrule
\multirow{4}{*}{TREC-6}  & SVM & \num{0.740+-0.000} & \num{0.758+-0.000} & \num{0.692+-0.101} & \num{0.596+-0.145} & \num{0.742+-0.031}\\
 & KimCNN & \num{0.840+-0.016} & \num{0.836+-0.012} & \num{0.834+-0.015} & \num{0.802+-0.017} & \num{0.792+-0.020}\\
 & DistilRoBERTa & \num{0.942+-0.008} & \num{0.950+-0.009} & \num{0.942+-0.009} & \num{0.940+-0.011} & \num{0.918+-0.016}\\
 & BERT & \num{0.932+-0.010} & \num{0.947+-0.014} & \num{0.960+-0.006} & \bfseries\num{0.968+-0.004} & \num{0.921+-0.025}\\
\bottomrule
\end{tabular}
\caption{%
Final accuracy per dataset, model, and query strategy. We report the mean and standard deviation over five runs. The best result per dataset is printed in bold.}
\label{table-results-acc}
\end{table*}

\section{Technical Environment}
 
All experiments were conducted within a Python 3.8 environment. The system had CUDA 11.1 installed and was equipped with an {NVIDIA GeForce RTX 2080 Ti} (11GB VRAM). Computations for fine-tuning transformers and training KimCNN were performed on the GPU.

\section{Implementation Details}

Our experiments were built using well-known machine learning libraries: PyTorch\footnote{\url{https://pytorch.org/}, 1.8.0}, huggingface transformers\footnote{\url{https://github.com/huggingface/transformers}, 4.11.0}, scikit-learn\footnote{\url{https://scikit-learn.org/}, 0.24.0}, scipy\footnote{\url{https://www.scipy.org/}, 1.6.0}, and numpy\footnote{\url{https://numpy.org/}, 1.19.5}. For active learning and text classification, we used small-text\footnote{\url{https://github.com/webis-de/small-text}, 1.0.0a8} \citep{schroeder:2022}.

\section{Experiments}

Each experiment configuration represents a combination of model, dataset and query strategy, and has been run for five times. We used a class-balanced initial set to support the warm start of the first model for the imbalanced TREC-6 dataset, whose rarest class would otherwise only rarely be encountered if sampled randomly.

\subsection{Pre-Trained Models}

We fine-tuned DistilRoBERTa (\href{https://huggingface.co/bert-base-uncased}{distilroberta-base}) and BERT-large (\href{https://huggingface.co/bert-large-uncased}{bert-large-uncased}). Both of them are available via the \href{https://huggingface.co/models}{huggingface model repository}.

\begin{table}[!htbp]%
\centering%
\fontsize{9pt}{10pt}\selectfont%
\renewcommand{\tabcolsep}{7pt}%
\begin{tabular}[t]{@{}l@{\hspace{16pt}}r@{}}
\toprule
\bfseries Dataset & \bfseries Max. Seq. Length\\
\midrule
AGN & 60 \\
CR & 50   \\
MR & 60 \\
SUBJ & 50 \\
TREC & 40 \\
    \bottomrule
\end{tabular}
\caption{Hyperparameter settings for the maximum sequence length (as number of tokens) per dataset.}
\label{table-sequence-length}
\end{table}

\begin{table*}[!b]%
\vspace*{5pt}
\centering
\fontsize{8pt}{9pt}\selectfont%
\renewcommand{\tabcolsep}{12pt}%
\begin{tabular}{@{}ll@{\hspace{20pt}}lllll@{}}
\toprule
\textbf{Dataset} & \textbf{Model} & \multicolumn{5}{c}{\textbf{Query Strategy}}\\
\cmidrule{3-7} & & \multicolumn{1}{c}{\hspace*{-12pt}PE} & \multicolumn{1}{c}{BT} & \multicolumn{1}{c}{LC} & \multicolumn{1}{c}{CA} & \multicolumn{1}{c}{RS}\\
\midrule
 \multirow{4}{*}{AGN}  & SVM & \num{0.693+-0.000} & \num{0.705+-0.000} & \num{0.690+-0.011} & \num{0.458+-0.057} & \num{0.699+-0.012}\\
 & KimCNN & \num{0.753+-0.005} & \num{0.791+-0.013} & \num{0.739+-0.019} & \num{0.699+-0.022} & \num{0.810+-0.013}\\
 & DistilRoBERTa & \num{0.855+-0.018} & \bfseries\num{0.875+-0.007} & \num{0.852+-0.018} & \num{0.863+-0.020} & \num{0.855+-0.006}\\
 & BERT & \num{0.858+-0.015} & \num{0.872+-0.005} & \num{0.848+-0.018} & \num{0.864+-0.012} & \num{0.849+-0.007}\\
\midrule
\multirow{4}{*}{CR}  & SVM & \multicolumn{2}{c}{\num{0.717+-0.000}}  & \num{0.713+-0.009} & \num{0.695+-0.009} & \num{0.718+-0.007}\\
 & KimCNN & \multicolumn{2}{c}{\num{0.713+-0.015}}  & \num{0.717+-0.009} & \num{0.707+-0.004} & \num{0.705+-0.014}\\
 & DistilRoBERTa & \multicolumn{2}{c}{\num{0.874+-0.012}}  & \num{0.875+-0.008} & \num{0.853+-0.019} & \num{0.870+-0.010}\\
 & BERT & \multicolumn{2}{c}{\bfseries\num{0.877+-0.011}}  & \num{0.857+-0.016} & \num{0.866+-0.017} & \num{0.868+-0.008}\\
\midrule
\multirow{4}{*}{MR}  & SVM & \multicolumn{2}{c}{\num{0.612+-0.000}}  & \num{0.615+-0.012} & \num{0.584+-0.018} & \num{0.597+-0.004}\\
 & KimCNN & \multicolumn{2}{c}{\num{0.674+-0.009}}  & \num{0.683+-0.015} & \num{0.671+-0.009} & \num{0.677+-0.011}\\
 & DistilRoBERTa & \multicolumn{2}{c}{\num{0.784+-0.013}}  & \num{0.786+-0.026} & \num{0.785+-0.010} & \num{0.783+-0.007}\\
 & BERT & \multicolumn{2}{c}{\bfseries\num{0.833+-0.013}}  & \num{0.831+-0.012} & \num{0.817+-0.009} & \num{0.827+-0.006}\\
\midrule
\multirow{4}{*}{SUBJ}  & SVM & \multicolumn{2}{c}{\num{0.801+-0.000}}  & \num{0.802+-0.003} & \num{0.768+-0.008} & \num{0.797+-0.010}\\
 & KimCNN & \multicolumn{2}{c}{\num{0.859+-0.013}}  & \num{0.841+-0.007} & \num{0.838+-0.011} & \num{0.864+-0.008}\\
 & DistilRoBERTa & \multicolumn{2}{c}{\num{0.924+-0.006}}  & \num{0.925+-0.003} & \num{0.915+-0.015} & \num{0.902+-0.008}\\
 & BERT & \multicolumn{2}{c}{\num{0.939+-0.007}}  & \num{0.938+-0.016} & \bfseries\num{0.943+-0.005} & \num{0.933+-0.005}\\
\midrule
\multirow{4}{*}{TREC-6}  & SVM & \num{0.491+-0.000} & \num{0.648+-0.000} & \num{0.538+-0.085} & \num{0.462+-0.112} & \num{0.619+-0.026}\\
 & KimCNN & \num{0.711+-0.010} & \num{0.714+-0.009} & \num{0.683+-0.029} & \num{0.639+-0.025} & \num{0.688+-0.013}\\
 & DistilRoBERTa & \num{0.840+-0.023} & \num{0.864+-0.014} & \num{0.860+-0.013} & \num{0.842+-0.005} & \num{0.856+-0.020}\\
 & BERT & \num{0.789+-0.032} & \num{0.844+-0.013} & \num{0.858+-0.030} & \bfseries\num{0.868+-0.027} & \num{0.828+-0.018}\\
\bottomrule
\end{tabular}
\caption{%
Final AUC per dataset, model, and query strategy. We report the mean and standard deviation over five runs. The best result per dataset is printed in bold.}
\label{table-results-auc}
\end{table*}

\begin{table*}[!t]%
\centering
\fontsize{8pt}{9pt}\selectfont%
\renewcommand{\tabcolsep}{6.75pt}%
\begin{tabular}{@{}ll@{\hspace{10pt}} r @{${}\pm{}$} r r @{${}\pm{}$} r r @{${}\pm{}$} r r @{${}\pm{}$} r r @{${}\color{gray}\pm{}$} r @{}}
\toprule
\textbf{Dataset} & \textbf{Model} & \multicolumn{10}{c}{\textbf{Query Strategy}}\\
\cmidrule{3-12} & & \multicolumn{2}{c}{\hspace*{-6pt}PE} & \multicolumn{2}{c}{BT} & \multicolumn{2}{c}{LC} & \multicolumn{2}{c}{CA} & \multicolumn{2}{c}{\hspace*{4pt}RS}\\
\midrule
\multirow{4}{*}{AGN}  & SVM & 1.852 & 0.415 & 0.907 & 0.203 & \bfseries 0.432 & \bfseries 0.097 & 516.554 & 115.583 & \color{gray}0.001 & \color{gray}0.000\\
 & KimCNN & 7.264 & 1.626 & \bfseries 6.199 & \bfseries 1.389 & 10.256 & 2.359 & 481.758 & 142.013 & \color{gray}0.002 & \color{gray}0.000\\
 & DistilRoBERTa & 97.479 & 21.800 & 96.372 & 21.551 & \bfseries 87.398 & \bfseries  19.560 & 852.457 & 230.157 & \color{gray}0.002 & \color{gray}0.000\\
 & BERT & 528.884 & 118.347 & 503.454 & 112.583 & \bfseries 480.401 & \bfseries  107.422 & 1475.960 & 391.579 & \color{gray}0.002 & \color{gray}0.000\\
\midrule
\multirow{4}{*}{CR}  & SVM & 0.005 & 0.001 & 0.005 & 0.001 & \bfseries  0.003 & \bfseries  0.001 & 0.307 & 0.070 & \color{gray}0.000 & \color{gray}0.000\\
 & KimCNN & 0.184 & 0.042 & \bfseries 0.155 & \bfseries 0.035 & 0.163 & 0.036 & 0.705 & 0.189 & \color{gray}0.000 & \color{gray}0.000\\
 & DistilRoBERTa & 1.942 & 0.434 & 1.916 & 0.428 & \bfseries 1.912 & \bfseries 0.428 & 2.627 & 0.648 & \color{gray}0.000 & \color{gray}0.000\\
 & BERT & \bfseries 12.112 & \bfseries 2.709 & 12.374 & 2.767 & 12.427 & 2.780 & 12.750 & 2.852 & \color{gray}0.000 & \color{gray}0.000\\
\midrule
\multirow{4}{*}{MR}  & SVM & 0.014 & 0.003 & 0.014 & 0.003 & \bfseries 0.009 & \bfseries 0.002 & 1.889 & 0.425 & \color{gray}0.000 & \color{gray}0.000\\
 & KimCNN & 0.521 & 0.117 & \bfseries 0.436 & \bfseries 0.098 & 0.468 & 0.105 & 3.672 & 1.098 & \color{gray}0.000 & \color{gray}0.000\\
 & DistilRoBERTa & 7.558 & 1.691 & 7.481 & 1.673 & \bfseries 7.183 & \bfseries  1.627 & 12.303 & 3.293 & \color{gray}0.000 & \color{gray}0.000\\
 & BERT & \bfseries 41.428 & \bfseries 9.265 & 42.247 & 9.447 & 41.960 & 9.391 & 43.480 & 9.747 & \color{gray}0.000 & \color{gray}0.000\\
\midrule
\multirow{4}{*}{SUBJ}  & SVM & 0.014 & 0.003 & 0.013 & 0.003 & \bfseries 0.009 & \bfseries 0.002 & 1.969 & 0.444 & \color{gray}0.000 & \color{gray}0.000\\
 & KimCNN & 0.472 & 0.106 & \bfseries 0.409 & \bfseries 0.091 & 1.708 & 1.144 & 3.161 & 0.954 & \color{gray}0.000 & \color{gray}0.000\\
 & DistilRoBERTa & 5.219 & 1.167 & 5.153 & 1.153 & \bfseries 5.099 & \bfseries 1.140 & 10.508 & 2.885 & \color{gray}0.000 & \color{gray}0.000\\
 & BERT & \bfseries 31.332 & \bfseries 7.006 & 32.908 & 7.358 & 33.043 & 7.393 & 37.832 & 8.478 & \color{gray}0.000 & \color{gray}0.000\\
\midrule
\multirow{4}{*}{TREC-6}  & SVM & 0.085 & 0.019 & 0.042 & 0.009 & \bfseries 0.018 & \bfseries 0.004 & 0.609 & 0.138 & \color{gray}0.000 & \color{gray}0.000\\
 & KimCNN & 0.289 & 0.065 & \bfseries 0.248 & \bfseries 0.055 & 1.111 & 0.745 & 1.504 & 0.447 & \color{gray}0.000 & \color{gray}0.000\\
 & DistilRoBERTa & 2.934 & 0.656 & \bfseries 2.887 & \bfseries 0.646 & 3.239 & 1.473 & 4.691 & 1.271 & \color{gray}0.000 & \color{gray}0.000\\
 & BERT & 14.577 & 3.260 & \bfseries 14.539 & \bfseries 3.251 & 14.963 & 3.350 & 17.901 & 17.213 & \color{gray}0.000 & \color{gray}0.000\\
\bottomrule
\end{tabular}
\caption{%
Query time in seconds. We report the mean and standard deviation over five runs. The best result (with the lowest query time) per dataset and model is printed in bold.}
\label{table-runtimes}
\end{table*}

\subsection{Datasets}

Our experiments used datasets that are well-known benchmarks in text classification and active learning. All datasets have been made accessible to the Python ecosystem by several Python libraries that provide fast access to the raw text of those datasets. We obtain CR and SUBJ using \href{https://nlp.gluon.ai}{gluonnlp}, and  AGN, MR, and TREC using \href{https://github.com/huggingface/datasets}{huggingface datasets}.

\subsection{Hyperparameters}

\paragraph{Maximum Sequence Lenght}

We set the maximum sequence length to the minimum multiple of ten for which 95\% of the respective dataset's sentences contain less than or an equal number of tokens for both KimCNN and transformers (shown in Table~\ref{table-sequence-length}).

\paragraph{Transformers}

AGN is trained for $50$~epochs and all other datasets for~$15$ epochs \citep{howard:2018}. For training, we use AdamW \citep{loshchilov:2019} with a learning rate of $\eta$~=~$\num{2e-5}$, beta coefficients of $\beta_1$~=~$0.9$ and $\beta_2$~=~$0.999$, and an epsilon of $\epsilon$~=~$\num{1e-8}$. Training is done in batches, with a batch size of $12$. 

\paragraph{KimCNN}

We adopt the parameters by \citet{zhang:2017}, i.e., $50$ filters and filter heights of $(3, 4, 5)$. Training is done in batches with a batch size of 25, a learning rate of $\eta$~=~$\num{1e-3}$, and word embeddings from word2vec \citep{mikolov:2013}.

\section{Standard Deviations and Runtimes}

In Table~\ref{table-results-acc} and Table~\ref{table-results-auc} we report final accuracy and AUC scores including standard deviations, measured after the last iteration of active learning. Moreover, we report the runtimes of the query step per strategy in Table~\ref{table-runtimes}.

\subsection{Evaluation Metrics}

Active learning was evaluated using standard active learning metrics, namely accuracy und area under the learning curve. For both metrics, the  respective scikit-learn implementation was used.

\putbib
\end{bibunit}

\end{document}